\documentclass[runningheads]{llncs}
\usepackage[T1]{fontenc}
\usepackage{graphicx}
\usepackage{hyperref}
\usepackage{url}
\usepackage{cite}
\usepackage{soul}
\usepackage{color}
\usepackage{subfig}
\usepackage{amssymb, amsmath}

\usepackage{color}

\begin{document}
\title{Super-Resolution Analysis for \\ Landfill Waste Classification}
\titlerunning{SR Analysis for Landfill Waste Classification.}
%
\author{Mat\'{i}as Molina\inst{1}\orcidID{0000-0001-8799-6725}
\and
Rita P. Ribeiro\inst{1,2}\orcidID{0000-0002-6852-8077}
\and
Bruno Veloso\inst{1,3}\orcidID{0000-0001-7980-0972}
\and
Jo\~{a}o Gama\inst{1,3}\orcidID{0000-0003-3357-1195}
}
\authorrunning{M. Molina et al.}
%
\institute{INESC TEC, Porto, Portugal. \\           
\email{matias.d.molina@inesctec.pt}
  \and 
 Faculty of Sciences, University of Porto, 4169-007 Porto, Portugal \\
  \email{rpribeiro@fc.up.pt}
            \and
Faculty of Economics, University of Porto, 4200-464 Porto. Portugal \\
\email{\{bveloso,jgama\}@fep.up.pt}}
\maketitle              
\begin{abstract}
Illegal landfills are a critical issue due to their environmental, economic, and public health impacts. This study leverages aerial imagery for environmental crime monitoring. While advances in artificial intelligence and computer vision hold promise, the challenge lies in training models with high-resolution literature datasets and adapting them to open-access low-resolution images. Considering the substantial quality differences and limited annotation, this research explores the adaptability of models across these domains. Motivated by the necessity for a comprehensive evaluation of waste detection algorithms, it advocates cross-domain classification and super-resolution enhancement to analyze the impact of different image resolutions on waste classification as an evaluation to combat the proliferation of illegal landfills. We observed performance improvements by enhancing image quality but noted an influence on model sensitivity, necessitating careful threshold fine-tuning.

\end{abstract}

\setcounter{footnote}{0} 
\section{Introduction}
Illegal landfills are places where waste material is dumped, violating management laws. Illegal dumping has a tremendous impact on our ecosystem, affecting our economy and health. As a result, various governments and institutions worldwide have invested additional resources to prevent the proliferation of waste dumping \cite{mager2022illegal, rodriguez2015overview, europol_2022_online}.

Aerial images from satellites, planes and, more recently, drones are an important source of information for earth observation that helps in fighting against environmental crime~\cite{esa_earthObservation, esa_satelliteTechnology}. In recent years, with the advance of artificial intelligence and particularly computer vision approaches due to the increasing computing capabilities and the exponential growth of available annotated data, it has become possible to build different and accurate models for image classification, as exemplified in numerous image recognition challenges, including the Large Scale Visual Recognition Challenge (LSVRC)~\cite{deng2009imagenet}, especially following the introduction of Convolutional Neural Networks (CNNs)~\cite{NIPS2012_c399862d}. However, the need for an initial amount of images to train such models makes it difficult to apply in waste detection. To build robust machine learning methods for landfill classification, obtaining a significant amount of landfill locations and access to their aerial images is crucial. The number of aerial landfill datasets is limited, and they are commonly high-resolution images. However, the location of these images is not typically provided due to confidential information agreements. On the other hand, although it is possible to obtain free access to different European satellite images (e.g. Sentinel-2 via Copernicus~\footnote{Copernicus hub: \url{https://scihub.copernicus.eu/}}), these images are typically of very low resolution.
In this context, we face two challenges: i) The classification models are based on high-resolution literature datasets with no geolocation provided, and they only contain the standard Red, Green, and Blue (RGB) bands. ii) The open-access satellite image banks provide low-resolution information, but we need information about illegal landfill locations.
These facts encourage us to investigate the adaptability of the classification models built on the literature dataset to different resolution query domains, as in the case of the freely available image banks. However, the tremendous quality difference between the two types of images and the lack of annotations present a hard limitation. 
For a comprehensive evaluative process on the road of building machine learning and computer vision waste detection algorithms, we motivate the evaluation of cross-domain classification together with super-resolution enhancement, which is the main goal of our work.

The main contributions of this manuscript lie in leveraging a dataset with exceptionally high resolution to construct a binary classification model and systematically evaluate the model's performance across various resolutions. Furthermore, we suggest an assessment of the model's capabilities when trained with higher-resolution images and subsequently applied to downscaled samples. Following these experiments, we advocate utilising a super-resolution model to enhance the quality before the classification process.

\section{Related Work}
This section explores two essential aspects: literature datasets used for landfill classification tasks and methodologies employed to enhance image resolution. Each subsection provides crucial insights into these key components of our study.

\subsection{Image classification for landfills discovery}
\label{section:aerialwaste}
The high performance of deep learning (DL) in image classification, particularly in identifying landfills in aerial photographs, relies on the availability of high-quality datasets to build the models. Unfortunately, such datasets have been lacking in the field, posing a significant challenge in developing scalable and accurate methods. To address this critical gap, Torres and Fraternali~\cite{torres2023aerialwaste} introduced the AerialWaste dataset for the specific task of landfill detection.
What sets this dataset apart is that it has been meticulously annotated by professional photo interpreters, making it valuable for building well-accurate image classifiers.
AerialWaste contains 10434 images of very-high-resolution images generated from different sources: AGEA Orthophotos (1000$\times$1000 pixels and 0.2m Ground Sampling Distance, GSD), WORLDView-3 (700$\times$700 pixels and 0.3m GSD) and GoogleEarth (1000$\times$1000 pixels and 0.5m GSD). They also provide 3478 positive and 6956 negative samples, indicating the presence or absence of waste, respectively.
The authors also provide the results of training a deep residual model for binary classification using the ResNet-50~\cite{he2016deep} architecture, initialized with transfer learning from ImageNet~\cite{deng2009imagenet} and augmented with a Feature Pyramid Network (FPN)~\cite{lin2017feature}.
Another deep learning model applied to landfill monitoring is RetinaNet~\cite{lin2017focal} with DenseNet~\cite{huang2017densely} as the backbone to identify landfills as an object detection task in satellite images of the Shanghai district~\cite{abdukhamet2019landfill, rs13224520}. One important conclusion of this study is the positive impact of data augmentation.

Despite the different literature models to address classification, detection or segmentation for landfills, in most cases, the primary challenge involves the lack of annotations and the different sizes of the contained waste. The recent public Aerial Waste dataset provides an important amount of annotated images and the evaluation of the binary classification through ResNet-50+FPN. We use this dataset for our experiments, and the ResNet model is the base for our classification benchmark.

\subsection{Image quality improvement}
Super-resolution (SR) is one of the most popular techniques that aim to enhance the resolution and quality of images. It is particularly useful when dealing with low-resolution images. The general idea behind SR is to learn to generate missing details in low-resolution images by correlating them with their original high-resolution pair.
Various models have been explored in the literature to address super-resolution for different tasks~\cite{ledig2017photo, nasrollahi2014super}. The standard optimization approach minimizes the mean squared error (MSE) between the original high-resolution image and the high-resolution version constructed from the low-resolution input. In addition, it is common to consider the convergence to the peak signal-to-noise ratio (PSNR) as a complementary measure. One of the most popular models uses a generative adversarial network (GAN) approach (SRGAN)~\cite{nasrollahi2014super} and employs a deep residual network (ResNet) architecture. Based on SRGAN,  Lim et al. propose an enhanced model (Enhanced Deep residual network for Super-Resolution, EDRS)~\cite{lim2017enhanced} as a simplified version of the SRGAN architecture by removing the unnecessary modules. The authors also suggest minimizing the L1 norm instead of the MSE or L2 norm. Such modification increases the original performance, improving the computation time. In the following sections, we explain our proposal based on using ResNet for classification and the EDSR baseline for our experiments.

In our work, we extend and incorporate the SR framework into the classification pipeline to address the specific challenges of landfill classification at different resolution settings. We perform an exhaustive evaluation to understand how super-resolution can serve as a pre-process, mitigating the resolution gap and enhancing the performance of our classification model.

\begin{figure}[!t]
\centering
\includegraphics[width=0.8\textwidth]{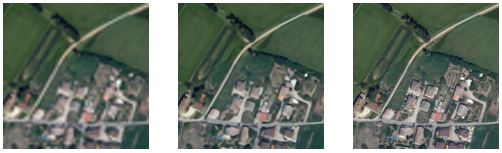}
\caption{Super-Resolution (SR) example: low-resolution input (left), SR output (center) and high-resolution ground truth (right).} \label{fig:SR_example}
\end{figure}

In Figure~\ref{fig:SR_example} it is possible to visualize an example of applying our implementation of the EDSR model on a downscaled sample of the Aerial Waste dataset.

\section{Methodology}
Since our study focuses on evaluating waste detection through image classification at different resolution domains, we leverage the advances in computer vision and deep learning for this task. We use high-resolution aerial images annotated with the presence or absence of waste.

We propose reproducing the ResNet-50 model with transfer learning from ImageNet and trained with aerial waste, as proposed in~\cite{torres2023aerialwaste}. To analyze the impact of the different scales, we are not augmenting the architecture with FPN.
The first experiment is to train the model by downscaling the entire dataset. It provides us with a general behaviour of the tasks throughout different resolutions. After this, we use the model trained with the highest resolution to query the different downscaled test samples to understand how much information is missing at each resolution domain. Finally, we propose using a super-resolution to improve each downscaled sample to the original image to analyze its impact when classifying.

In summary, the three experimental scenarios are:
\begin{itemize}
    \item[I.] Waste classification at different resolutions: ResNet-50 for binary classification at different resolutions given by the different image size dimensions \{256, 128, 64, 32, 16, 8\}.
    The output of this experiment aims to answer the following question: \textit{How good is the model performance while trained with data of inferior quality?}
    
    \item[II.] High-resolution training for waste classification: ResNet-50 for binary classification, train with high-resolution images and querying with different lower resolutions.
    The output of this experiment aims to answer the following question: \textit{How good is the model performance when it is queried with data of inferior quality?}

    \item[III.] Waste classification and super-resolution enhancement: ResNet-50 for binary classification and Enhanced Deep Super-Resolution network (EDSR) for image quality improvement. The EDSR is trained to be applied to images of different resolutions to match the resolution of the images on which the ResNet model has been trained.
    The output of this experiment aims to answer the following question: \textit{How good is the classification model when it is combined with a super-resolution enhancement as a pre-processing step?}
\end{itemize}

In all the cases, we use the provided train and test split and also split the training part to use the $20\%$ for validation. We found the best learning rate by a grid search into \{1e-2, 1e-3, 1e-4, 1e-5\}, with 1e-4 being the best value. Moreover, as the ResNet is pre-trained with ImageNet, we did an initial validation, freezing a different number of layers to find the best possible option to avoid overfitting. Freezing the first two layers best works for us (as also worked in \cite{torres2023aerialwaste}).

\subsection{Experimental Setup}

\subsubsection{Metrics.} We evaluate the classification model with the standard Accuracy, Precision, Recall and F1 Score metrics for a comprehensive analysis. In addition, we also consider the True Positive Rate (TPR) and False Positive Rate (FPR) for a deeper analysis, as is discussed in Section~\ref{section:results}.

For super-resolution, we use the two widely used metrics, Peak Signal-to-Noise Ratio (PSNR) and Structural Similarity Index (SSIM)~\cite{lim2017enhanced}.

\subsubsection{Experiment I.}

\begin{table}[b]
\caption{Different downscales and respective resolution (meters) for each source.}\label{table:downscaling}
\centering
\begin{tabular}{|c|c|c|c|c|c|}
\hline
Downscale       &  Size (px)  & AGEA (m) & WV3 (m) & Google (m)      \\
\hline
0 (initial)& 256 & 0.78 & 0.82 & 1.95 \\
1 ($1/2$)  & 128 & 1.56 & 1.64 & 3.9  \\
2 ($1/4$)  & 64  & 3.10 & 3.28 & 7.8  \\
3 ($1/8$)  & 32  & 6.24 & 6.56 & 15.6 \\
4 ($1/16$) & 16  & 12.48 & 26.24 & 31.2 \\
5 ($1/32$) & 8   & 24.96 & 52.48 & 62.4 \\
\hline
\end{tabular}
\end{table}

The initial experiment involves training ResNet classifiers on different dataset resolutions. Initially, the dataset is resized to $512\times512$, and subsequent experiments with a size of $256\times256$ yielded similar performance. Therefore, $256\times256$ is chosen as the initial size for further experiments, representing the highest resolution. 

The experiment outputs ResNet-50 models trained on images of sizes $256\times256$, $128\times128$, $64\times64$, $32\times32$, $16\times16$, and $8\times8$. This corresponds to scaling the images down by factors of $1/2^5$ ($1/2$, $1/4$, $1/8$, $1/16$, and $1/32$, respectively). Considering the original dataset size and resolutions, as described in Section~\ref{section:aerialwaste}, three different resolutions are obtained for each image source, and the resulting resolutions are calculated with respect to each resizing. These details are summarized in Table~\ref{table:downscaling}.
Despite the original dataset having various sizes and resolutions, this table serves as a reference, emphasizing that beyond the size of $32\times32$, we start approaching the resolution of open and accessible satellite imagery banks.
We denote different models by indicating the input size in a sub-index:\
$ResNet_{256}$, $ResNet_{128}$, ..., $ResNet_{8}$.

\subsubsection{Experiment II.}
The goal of the second experiment is to analyze the behaviour of the model trained with our higher size ($ResNet_{256}$)  and apply it across all different resolutions (sizes of 128, 64, 32, 16 and 8).
In this experiment, the model is trained only once using 256-size images. After that, it queries using the same testing split with different image sizes.

\subsubsection{Experiment III.}
The final experiment aims to enhance the quality of the downscaled images to improve the final performance of the classification task. We use super-resolution to enhance the different downscaled images to the original input size of 256 and then classify them with the same ResNet model trained with images of size 256. We implement other EDSR models to learn how to improve each downscaled image to the original size of 256~$\times$~256.

In this case, we trained only one model for classification $ResNet_{256}$, and we built different super-resolution models $SR_{128\rightarrow256}, SR_{64\rightarrow 256},.., SR_{8\rightarrow 256}$.

\subsubsection{Training and evaluation.}
The ResNet model was trained using the original split provided by the dataset authors. The validation was performed using the $20\%$ of the training set. We stop the training at the epoch in which the loss training function converges, and the validation loss function deviates from it. We use the ResNet-50 architecture and the ImageNet pre-trained options. Depending on the input size setting, we needed between 17 and 20 epochs to train the model. Despite that, the process behaviour is similar in all the cases.

The Super Resolution model (EDSR) was trained using the dataset resize to $256\times256$ as high-resolution and each of one of its downscaled versions as low-resolution. Continuing the high- and low-resolution input pairs. In this case, we follow three different metrics: the loss function (L1 loss), Pick Signal to Noise Radio (PSNR) and the structural similarity index measure (SSMI). We focused on the loss and PSNR curves more than the SSMI since the different downscaled images make it more challenging to achieve a good SSMI even though it also converges.

\subsection{Results}
\label{section:results}
\begin{figure}[!hbt]
\centering
\includegraphics[width=0.62\textwidth]{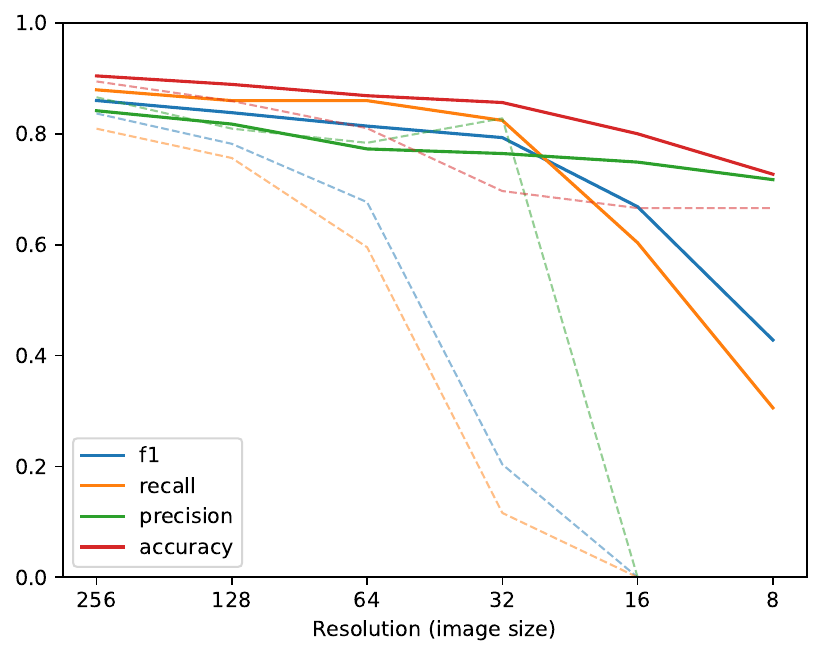}
\caption{
Evaluation of models trained on different resolutions (Experiment I) - solid lines - and on maximum resolution (Experiment II) - dashed lines - by image size.} \label{fig:experiment1-2}
\end{figure}

The results of experiments I and II, involving training and testing the model at different resolutions, are summarized in Table~\ref{table:experiment1-2}. For a more comprehensive comparison, refer to Figure~\ref{fig:experiment1-2}, where solid lines depict the different metrics of training and testing the classification model at each resolution. While the model shows reasonable performance across different resolutions, a noticeable decline occurs beyond a certain point, especially when the resolution is downscaled to size 32. Dashed lines represent the performance of training the model with high-resolution images (size 256) and testing with each lower resolution. An important observation is that despite is possible to obtain a suitable performance at lower resolutions, from a certain point in advance, the performance declines with a stronger tendency  (e.g., training with a model for image size 256 is significantly less effective for images of size 32, 16, or 8). This suggests that (taking our reference in Table~\ref{table:downscaling}) training a model for a resolution of 1 meter does not ensure good performance at 10 meters.
We demonstrate quantitatively that these observations align with the literature concerning the lack of annotation and the resolution gap. 
These challenges hinder the classification model adaptation. 

\begin{table}[!hbt]
\caption{
Evaluation of models trained on different resolutions (Experiment I) and maximum resolution (Experiment II) by image size.
}\label{table:experiment1-2}
\centering
\begin{tabular}{|c|c|c|c|c||c|c|c|c|}
\hline
Size & \multicolumn{4}{c||}{Experiment I} & \multicolumn{4}{c|}{Experiment II} \\
\cline{2-9}
 (px) & F1-score & Recall & Precision & Accuracy   & F1-score & Recall & Precision & Accuracy \\
\hline
\textbf{256} & 0.8600 & 0.8793 & 0.8416 & 0.9044   & 0.8366 & 0.8092 & 0.8659 & 0.8944 \\
\textbf{128} & 0.8381 & 0.8598 & 0.8175 & 0.8891   & 0.7819 & 0.7563 & 0.8093 & 0.8591 \\
\textbf{64}  & 0.8139 & 0.8598 & 0.7727 & 0.8687   & 0.6767 & 0.5954 & 0.7837 & 0.8100 \\
\textbf{32}  & 0.7931 & 0.8241 & 0.7644 & 0.8564   & 0.2036 & 0.1161 & 0.8279 & 0.6967 \\
\textbf{16}  & 0.6684 & 0.6034 & 0.7489 & 0.8000   & 0.0000 & 0.0000 & 0.0000 & 0.6660 \\
\textbf{8}   & 0.4280 & 0.3057 & 0.7173 & 0.7271   & 0.0000 & 0.0000 & 0.0000 & 0.6660 \\
\hline
\end{tabular}
\end{table}

Experiment III is summarized in Table~\ref{table:experiment3_SR_to256_ResNet_im256}. It shows the results of the super-resolution enhancement by setting a standard classification threshold at $0.5$ compared to the results after choosing the best possible threshold.

\begin{figure}[!bt]
\centering
\includegraphics[width=0.62\textwidth]{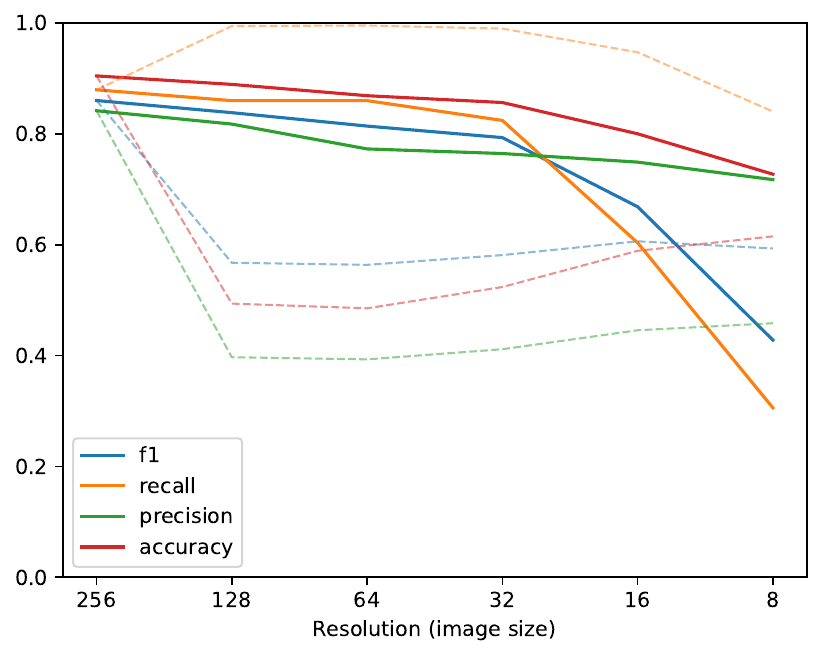}
\caption{Evaluation of models  
trained on different resolutions (Experiment I) - solid lines - and SR enhancement with default threshold (Experiment III) - dashed lines - by image size.
}
\label{fig:experiment1-3}
\end{figure}

\begin{figure}[!bt]
\centering
\includegraphics[width=0.62\textwidth]{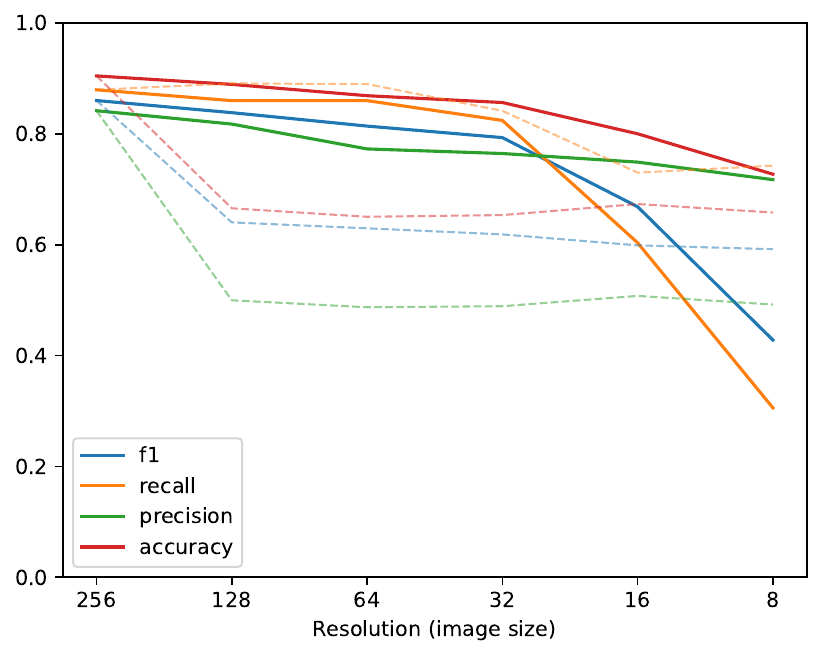}
\caption{Evaluation of models  
trained on different resolutions (Experiment I) - solid lines - and SR enhancement with best threshold (Experiment III) - dashed lines - by image size.
} \label{fig:experiment3_thr}
\end{figure}

In Figure~\ref{fig:experiment1-3}, we observe the impact of using super-resolution (SR) to improve the quality when querying the model trained with higher-resolution images. The final result shows intermediate performance between Experiment I and Experiment II, which means the SR approach improves when querying with lower resolutions than the training. Nevertheless, it is evident how this methodology impacts the model sensitivity (high recall) that biases the F1 Score up. To better understand the behaviour of the model sensitivity when SR is applied, we analyzed the true positive rate and the false positive rate throughout different threshold values during testing. 
The impact of changing the default classification threshold during testing could be examined in Figure~\ref{fig:experiment3_thr}. The plot shows a model with lesser sensitivity than the default threshold while maintaining the F1 score with no significant decrease.

\begin{table}[!hbt]
\caption{
Evaluation of the model with Super-Resolution enhancement on downscaled images (Experiment III): default vs best threshold by image size.
}\label{table:experiment3_SR_to256_ResNet_im256}
\centering
\begin{tabular}{|c|c|c|c|c||c|c|c|c|}
\hline
Size & \multicolumn{4}{c||}{Default threshold} & \multicolumn{4}{c|}{Best threshold} 
\\
 \cline{2-9}
 (px) & F1-score & Recall & Precision & Accuracy   & F1-score & Recall & Precision & Accuracy \\
\hline
\textbf{256} & 0.8600 & 0.8793 & 0.8416 & 0.9044 & 0.8600 & 0.8793 & 0.8416 & 0.9044 \\
\textbf{128} & 0.5674 & 0.9943 & 0.3970 & 0.4937 & 0.6402 & 0.8908$\blacktriangledown$ & 0.4997$\blacktriangle$ & 0.6656 \\
\textbf{64} & 0.5636 & 0.9954 & 0.3931 & 0.4852 & 0.6295 & 0.8897$\blacktriangledown$ & 0.4871$\blacktriangle$ & 0.6503 \\
\textbf{32} & 0.5812 & 0.9897 & 0.4114 & 0.5236 & 0.6185 & 0.8414$\blacktriangledown$ & 0.4890$\blacktriangle$ & 0.6534 \\
\textbf{16} & 0.6061 & 0.9471 & 0.4456 & 0.5889 & 0.5988 & 0.7299$\blacktriangledown$ & 0.5076$\blacktriangle$ & 0.6733 \\
\textbf{8} & 0.5931 & 0.8402 & 0.4583 & 0.6150 & 0.5918 & 0.7425$\blacktriangledown$ & 0.4920$\blacktriangle$ & 0.6580 \\
\hline
\end{tabular}
\end{table}

\begin{table}[!hbt]
\caption{Best thresholds for SR enhancement (Experiment III) based on the ROC curves by image size.}\label{table:threshold}
\centering
\begin{tabular}{|c|c|c|c|c|c|}
\hline
 Size      & \textbf{128} & \textbf{64} & \textbf{32} & \textbf{16} & \textbf{8} \\
 \hline
 Threshold & 0.97 & 0.98 & 0.97 & 0.92 & 0.76 \\
\hline
\end{tabular}
\end{table}

\begin{figure}[!hbt]
\centering
\subfloat[$ResNet \circ SR_{8\rightarrow 256}$] {\includegraphics[width=0.47\textwidth]{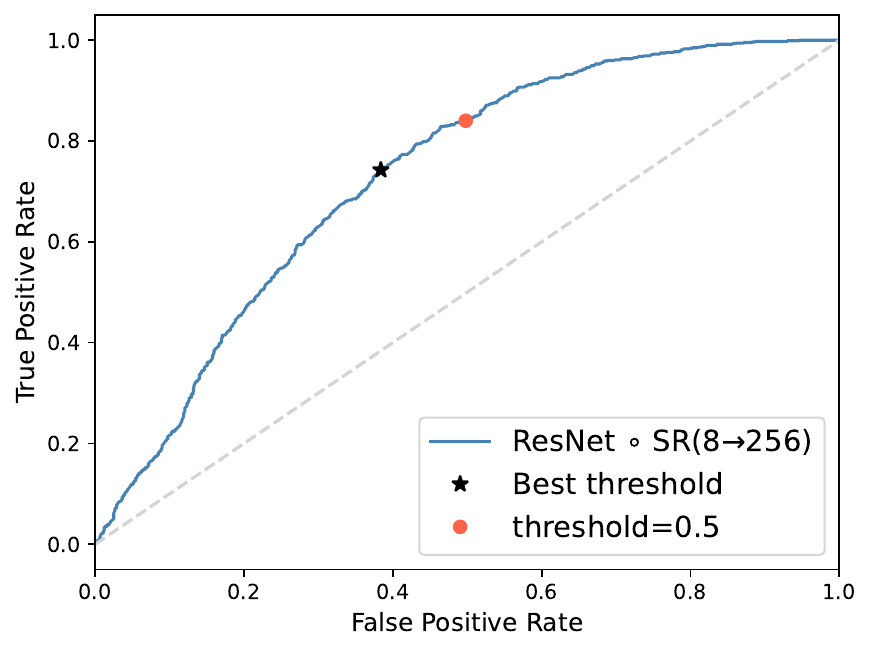}}
\subfloat[$ResNet \circ SR_{16\rightarrow 256}$] {\includegraphics[width=0.47\textwidth]{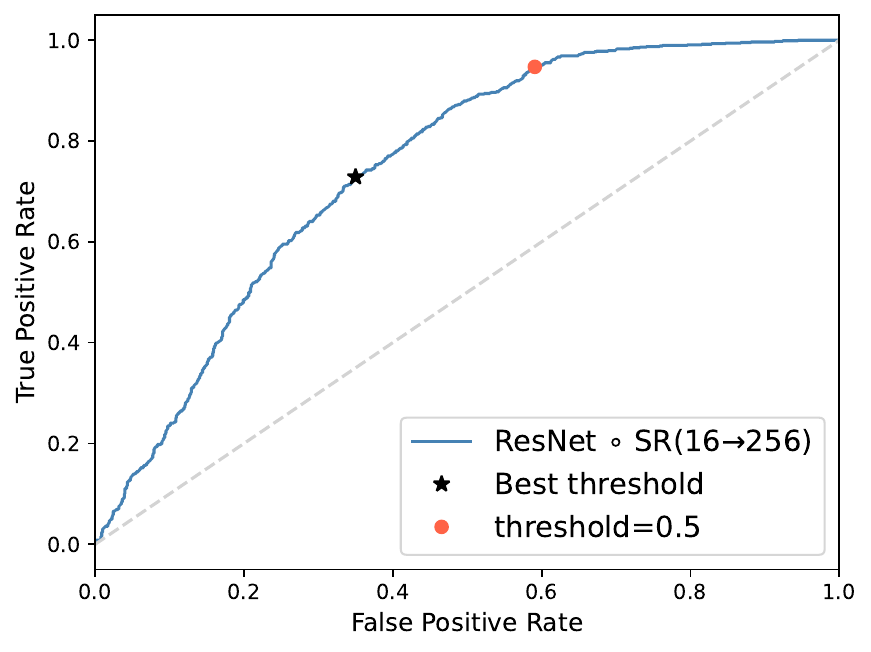}} \\
\subfloat[$ResNet \circ SR_{32\rightarrow 256}$] {\includegraphics[width=0.47\textwidth]{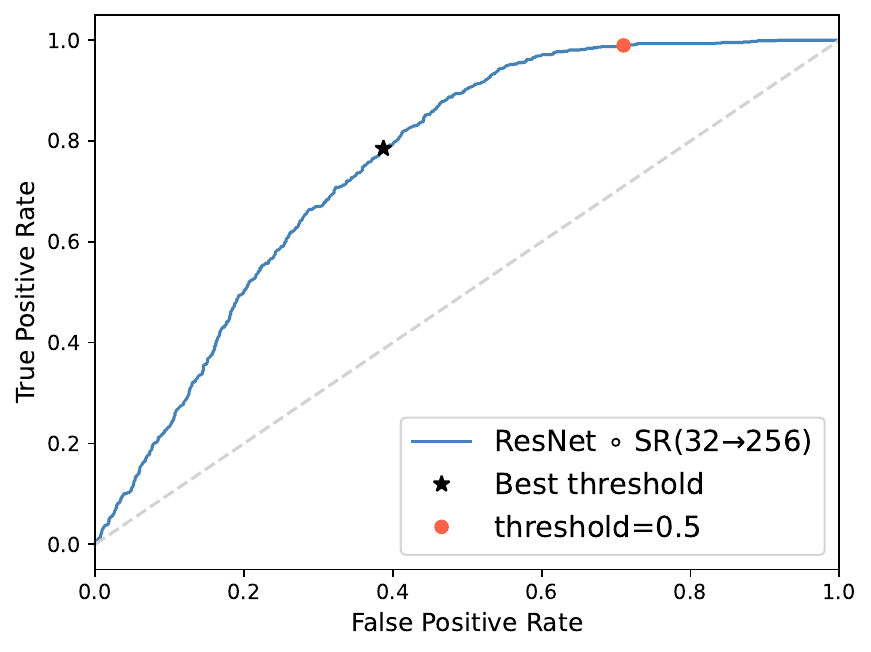}}
\subfloat[$ResNet \circ SR_{64\rightarrow 256}$] {\includegraphics[width=0.47\textwidth]{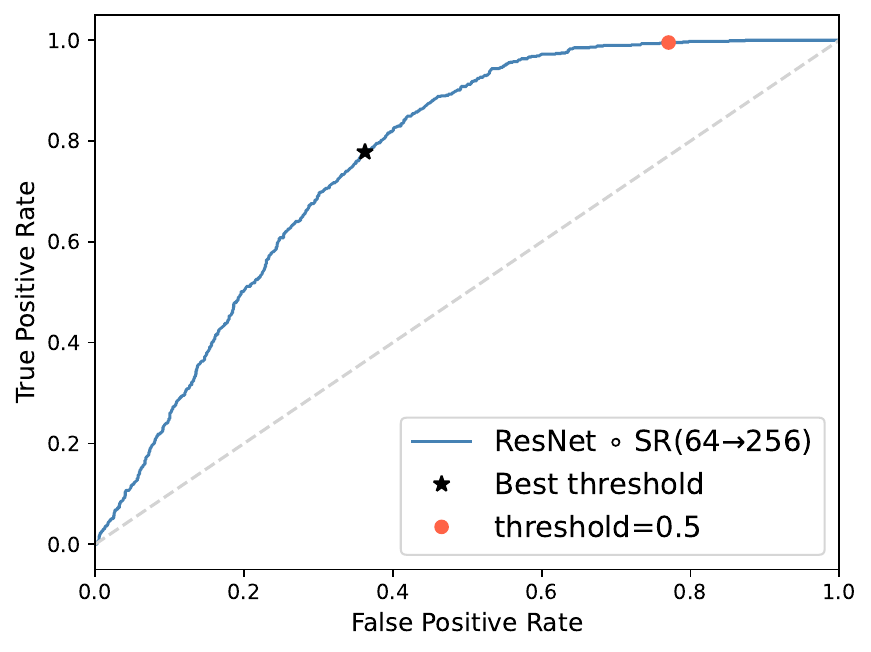}} \\
\subfloat[$ResNet \circ SR_{128\rightarrow 256}$] {\includegraphics[width=0.47\textwidth]{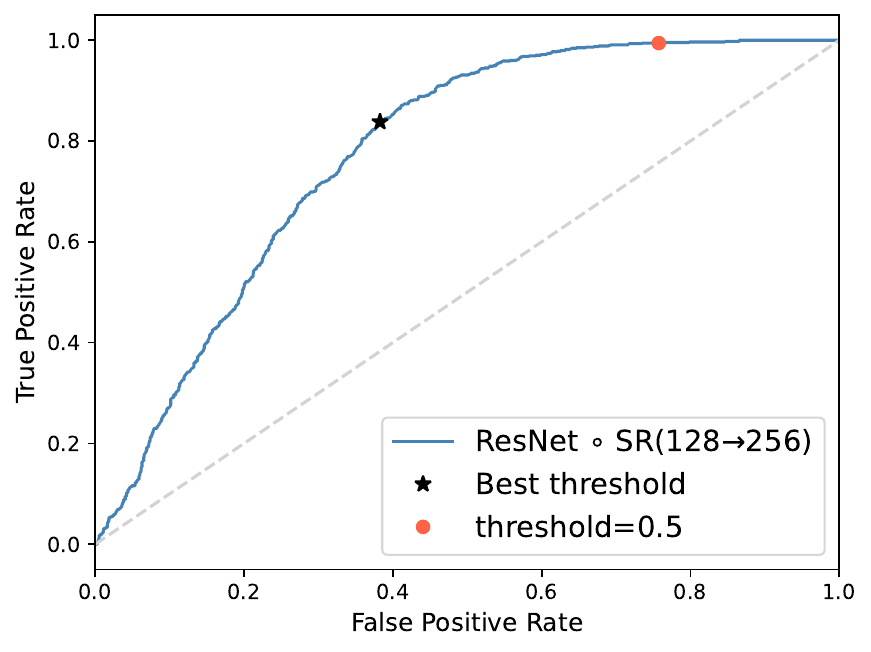}}
\caption{ROC curves obtained with SR enhancement (Experiment III). 
}
\label{fig:rocs}
\end{figure}

Regarding the threshold selection process, we could refer to the different Receiver Operating Characteristic curves (ROC) shown in Figure~\ref{fig:rocs}.
The ROC curves indicate high values for both TPR and FPR rates when the threshold is set to $0.5$ (shown as a red circle). The best threshold, thus, that makes a good compromise between both rates is plotted as a black star. It was revealed that higher thresholds can lead to a better compromise between both rates (Table~\ref{table:threshold}. We consider a $95\%$ as enough confidence level, and consequently, we set to $0.95$ all the thresholds greater than that. 
Another interesting observation is that the threshold maximizing both rates decreases as the original image resolution is reduced. These thresholds converge to the default one as the resolution decreases. We can conclude that the model's sensitivity can be adjusted, and its utility depends on the specific problem domain. A more sensitive model may be preferred in environmental crimes, particularly in detecting illegal landfills. This allows environmental agencies and governments to verify potential issues on-site. However, the model might incur a higher false negative rate in other scenarios, potentially missing a possible illegal landfill. This study motivates further exploration of various domain applications to provide a more comprehensive and cross-functional evaluation.

In summary, super-resolution enhances classification performance when the resolution of the query image is lower than the resolutions used during training. However, it also increases the model's sensitivity. As a result, it becomes essential to fine-tune the threshold according to the domain's specific requirements.


\section{Conclusions}
This study makes contributions by introducing a dual-model approach, combining a classification model and a super-resolution model, to address challenges in waste detection related to varying image resolutions. The methodology focused on combining both models to analyze the performance of waste classification and the pertinence of resolution enhancement motivated by the lack of information in conjunction with the different resolutions, which is a crucial challenge in literature. The research design involved a series of experiments to fairly compare image quality's influence on building classification models and explore potential improvements.
The experiments reveal important insights into the performance of waste detection models at various image resolutions. It evidences that the classifiers can be effectively trained at different resolutions. Still, their effectiveness declines as the resolution decreases, especially after the third downscaling, which resolutions correlate with the low resolution provided by the various open-access satellite imagery services. While training with high-resolution images initially yields favourable results, performance deteriorates as the resolution decreases. This indicates that a model designed for higher resolutions performs poorly at lower resolutions, especially after the first downscaling. These observations align with the challenges highlighted in existing literature, demonstrating the impact of the scarcity of annotations and the resolution gap.

Furthermore, we observe improvement in terms of performance when applying super-resolution to enhance image quality. This approach notably impacts the model's sensitivity, leading to higher recall and, consequently, increased F1 Score. The study also explores the model's sensitivity at varying threshold values, showcasing the importance of adjusting thresholds to match specific problem domains. In summary, super-resolution improves the waste classification performance for lower-resolution images. Still, it intensifies the model's sensitivity, underscoring the need for threshold fine-tuning tailored to specific domain requirements.

For future research, we intend to assess different super-resolution models and verify the variability using various public and collected datasets.

\section*{\ackname} 
This work was supported by the EMERITUS project, which was funded by the European Union's Horizon Europe research and innovation programme under Grant Agreement 101073874.



\bibliographystyle{splncs04}

\end{document}